\documentclass[runningheads]{llncs}
\usepackage{amsmath,graphicx}
\usepackage{float}
\usepackage[caption = false]{subfig}
\usepackage{algorithm}
\usepackage[noend]{algpseudocode}
\usepackage[numbers,sort&compress]{natbib}
\usepackage{booktabs,siunitx}
\begin{document}
\title{A Black-Box Attack on Optical Character Recognition Systems}
%
\author{Samet Bayram\inst{1}\orcidID{0000-0002-5821-4920} \and 
Kenneth Barner\inst{2}\orcidID{0000-0002-0936-7840}}
\authorrunning{S. Bayram and K. Barner}
%
\institute{University of Delaware, Newark De 19716, USA \email{sbayram@udel.edu} \and
University of Delaware, Newark De 19716, USA \email{barner@udel.edu}}

\maketitle              
\begin{abstract}
Adversarial machine learning is an emerging area showing
the vulnerability of deep learning models. Exploring attack methods to challenge state–of–the–art artificial intelligence (AI) models is an area of critical concern. The reliability and robustness of such AI models are one of the major concerns with an increasing number of effective adversarial attack methods. Classification tasks are a major vulnerable area for adversarial attacks. The majority of attack strategies are developed for colored or gray--scaled images. Consequently, adversarial attacks on binary image recognition systems have not been sufficiently studied. Binary images are simple --- two possible pixel-valued signals with a single channel. The simplicity of binary images has a significant advantage compared to colored and gray scaled images, namely computation efficiency. Moreover, most optical character recognition systems (OCRs), such as handwritten character recognition, plate number identification, and bank check recognition systems, use binary images or binarization in their processing steps. In this paper, we propose a simple yet efficient attack method, Efficient Combinatorial Black-box Adversarial Attack (ECoBA), on binary image classifiers. We validate the efficiency of the attack technique on two different data sets and three classification networks, demonstrating its performance. Furthermore, we compare our proposed method with state-of-the-art methods regarding advantages and disadvantages as well as applicability. 

\end{abstract}
\section{Introduction}

The existence of adversarial examples has drawn significant attention to the machine learning community. Showing the vulnerabilities of machine learning algorithms has opened critical research areas on the attack and robustness areas. Studies have shown that Adversarial attacks are highly effective on many existing AI systems, especially on image classification tasks \cite{dalvi,Biggio_2018,szegedy2014intriguing}. In recent years, a significant number of attack and defense algorithms were proposed for colored and gray--scaled images \cite{harnesing,carlini17,nguyen2015deep,moosavidezfooli2016deepfool,sharif,kurakin16,eykholt2018robust}. In contrast,  AI binary image adversarial attacks and defenses are not well studied. Existing attack algorithms are inefficient or not well suited to binary image classifiers because of the binary nature of such images. We explain the inefficiency of existing attack methods under the Related Works section.  

Binary image classification and recognition models are widely used in daily image processing tasks, such as license plate number recognizing, bank check processing, and fingerprint recognition systems. Critically, binarization is a pre--processing step for OCR systems, such as Tesseract \cite{tesseractsrev07}. The fundamental difference between binary and color/grayscale images, regarding generating adversarial examples, is their pixel value domains. Traditional color and grayscale attacks do not lend themselves to binary images because of their limited black/white pixel range. Specifically, color and grayscale images have a large range of pixel values, which allows crafting small perturbations to affect the desired (negative) classification result. Consequently, it is possible to generate imperceptive perturbations for color and grayscale images. However, in terms of perception, such results are much more challenging for binary images because there are only two options for the pixel values. Thus, a different approach is necessary to create attack methods for binary image classifiers. Moreover, the number of added, removed, or shifted pixels should be constrained to minimize the visual perception of attack perturbations. 

In this study, we introduce a simple yet efficient attack method in black--box settings for binary image classification models. Black--box attack only requires access to the classifier's input and output information. The presented results show the efficiency and performance of the attack method on different data sets as well as on multiple binary image classification models.

\subsection{Related Works}
Szegedy \emph{et al.} \cite{szegedy2014intriguing} show that even small perturbations in input testing images can significantly change the classification accuracy. Goodfellow \emph{et al.} \cite{harnesing} attempts to explain the existence of adversarial examples and proposes one of the first efficient attack algorithms in white--box settings. Madry \emph{et al.} \cite{madry2019deep} proposed projected gradient descent (PGD) as a universal first-order adversarial attack. They stated that the network architecture and capacity play a big role in adversarial robustness. One extreme case of an adversarial attack was proposed by \cite{onepixel}. In their study, they only changed the value of a single-pixel of an input image to mislead the classifier. Tramèr \emph{et al.} \cite{transfer} show the transferability of black-box attack among different ML models. Balkanski \emph{et al.} \cite{scar} proposes an attack method, referred to as scar, on binary image recognition systems. Scar resembles one of our perturbation models, namely additive perturbations. In this attack, it adds perturbation in the background of characters. Scar tries to hide the perturbations by placing them close to the character. However, this requires more perturbations to mislead the classifier. 

\subsubsection{Inefficiency of Previous Attack Methods }

\begin{figure}[H]
\centering
\subfloat[original image]{\includegraphics[width = .29\textwidth]{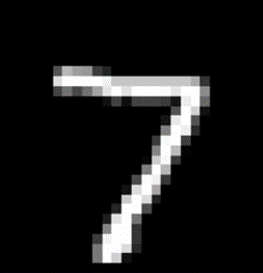}} 
\subfloat[perturbed image]{\includegraphics[width = .301\textwidth]{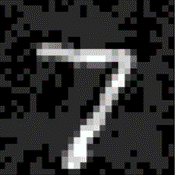}}
\subfloat[perturbed image]{\includegraphics[width = .304\textwidth]{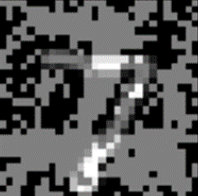}}\\
\subfloat[binary]{\includegraphics[width = .29\textwidth]{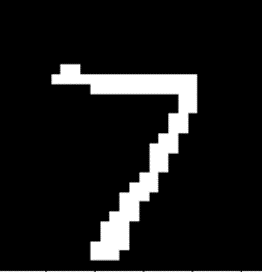}}
\subfloat[binary]{\includegraphics[width = .31\textwidth]{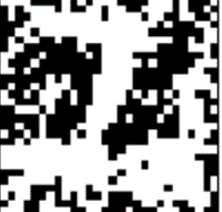}}
\caption{The effect of binarization on adversarial examples created by the PGD method. Perturbations in (b) are smaller than the binarization threshold, and perturbations in (c) are more significant than the threshold. (d) and (e) are the binary versions of (b) and (c), respectively. }
\label{fig:ineff}
\end{figure}

Attacking the binary classifiers should not be a complex problem at first sight. The attack method can only generate white or black pixels. However, having only two possible pixel values narrow downs the attack ideas. State-of-the-art methods such as PGD or FGSM create small perturbations to make adversarial examples look like the original input image. Those attack methods are inefficient on binary images because the binarization process wipes the attack perturbations in the adversarial example before it’s fed to the binary image classifier. This method, binarizing the input image, is considered a simple defense method against state-of-the-art adversarial attacks. Wang \emph{et al.} \cite{binarization_deffense} proposed a defense method against adversarial attacks by binarizing the input image as a pre-processing step before the classification. They achieved 91.2\% accuracy against white-box attacks on MNIST digits. To illustrate this phenomenon, we apply PGD on a gray scaled digit image whose ground truth label is seven. The PGD attack fools the gray-scaled digit classifier resulting output of three. However, after the binarization process of the same adversarial example, the perturbations generated by PGD are removed, and the image is classified as seven, as illustrated in Figure~\ref{fig:ineff}. For this reason, the state of art methods that generate perturbations less than the binarization threshold is inefficient when the image is converted to binary form. 
Adversarial perturbations created by the PGD attack method is disappeared when the perturbations are smaller than the binarization threshold. After the binarization process, the image is classified correctly for the case where perturbations are smaller than the threshold. On the other hand, adversarial perturbations can go through binarization when perturbations are bigger than the threshold. However, the final adversarial example contains an excessive amount of perturbations that ruin the main character (digit or letter), which is an unwanted situation for creating adversarial examples. Our proposed method produces a few white and black perturbations that can pass through the binarization process and misleads the classifiers. We show those perturbations in the Figure~\ref{fig:attack_tog}.

\section{PROBLEM DEFINITION}
\label{sec:format}

Let  $\boldsymbol{x}$  be a rasterized binary image with $d \times 1$ dimension. Each element of $\boldsymbol{x}$ is either 0 (black) or 1 (white). A trained multi--class binary image classifier $F$ takes $\boldsymbol{x}$ as input and gives $\boldsymbol{n}$ probabilities for each class. The label with highest probabilty, $y$, is the predicted label of $\boldsymbol{x}$. Thus, 
$ y = arg max_i F(\boldsymbol{x})_i$, where $F(\cdot )_i$ defines the binary image classifier, $i \in n$, and $n$ is the number of classes.

\subsection{Adversarial Example}
\label{ssec:adversarial}

An adversarial variation of $\boldsymbol{x}$ is $\boldsymbol{\tilde{x}}$, and its label denoted as $\tilde{y}$. Ideally, $\boldsymbol{\tilde{x}}$ resembles $\boldsymbol{x}$ as much as possible (metrically and/or perceptually), while $ y \neq \tilde{y} $. The classical mini--max optimization is adopted in this setting. That is, we want to maximize the similarity between the original input and adversarial example while minimizing the confidence of the true label $y$. While minimizing the confidence is generally straightforward, hiding the perturbations in adversarial samples is challenging in the binary image case, especially in low (spatial) resolution images.

\section{PROPOSED METHOD}
\label{sec:proposed methods}

Here we propose a black--box adversarial attack on binary image classifiers. The ECoBA consists of two important components: additive perturbations and erosive perturbations. We separate perturbations into two categories to have full control over whether to apply the perturbations to the character. The additive perturbations occur in the character's background, while the erosive perturbations appear on the character. Since preventing the visibility of attack perturbations is impossible for the binary image case, it is important to damage the character as less as possible while fooling the classifier successfully. Since the images are binary, we assume, without loss of generality, that white pixels represent the characters in the image, and black pixels represent the background. Proposed attack algorithms change the pixel value based on the decline in classification accuracy. We define this change as adversarial error, $\epsilon_i$, for the flipped $i^{th}$ pixel of input $\boldsymbol{x}$. For instance, $\boldsymbol{x} + \boldsymbol{w}_i $ means image $\boldsymbol{x}$ with $i^{th}$ pixel is flipped, from black to white. Thus, the adversarial example is $\boldsymbol{\tilde{x}} = \boldsymbol{x} + \boldsymbol{w}_i$, and the adversarial error is simply $\epsilon_i = \boldsymbol{x}_i - \boldsymbol{\tilde{x}_i}$, for the flipped $i^{th}$ pixel.

\subsection{Additive Perturbations}
\label{ssec:aca}

To create additive perturbations, an image is scanned, flipping each background (black) pixel, one by one, in an exhaustive fashion. The performance of the classifier is recorded for each potential pixel flip, and the results are ordered and saved in a dictionary, $D_{AP}$,  with the corresponding pixel index that causes the error. Pixels switched from black to white are denoted as $\boldsymbol{w}_i$, where $i$ represents the pixel index. The procedure is repeated, with $k$ indicating the number of flipped pixels, starting with the highest error in the dictionary and continuing until the desired performance level or the number of flipped pixels is achieved. That is, notionally
\begin{equation}
\arg \min_i F( \boldsymbol{x} + \boldsymbol{w}_i)  \text{ where } \| \boldsymbol{\tilde{x}} - \boldsymbol{x}\|_0 \le k . 
\end{equation}

The confidence of the adversarial example applied to the classifier is recorded after each iteration. If $\epsilon_i>0 $, then the $i^{th}$ pixel index is saved. Otherwise, the procedure is repeated, skipping to the next pixel. The procedure is completed by considering each pixel in the image.

\subsection{Erosive Perturbations}
\label{ssec:eca}

In contrast to additive perturbations, creating erosive perturbations is the mirror procedure. That is, pixels on the character (white pixels) are identified that cause the most significant adversarial error and thus flipped. Although previous works \cite{scar} utilize perturbing around or on the border of the character in an input image, erosive perturbations occur directly on (or within) the character. This approach can provide some advantages regarding the visibility of perturbations and maximizes the similarity between the original image and its adversarial example. Similarly, the sorted errors are saved in a dictionary, $D_{EP}$,  with the corresponding pixel index that causes the error. Pixels flipped from white to black are denoted as $\boldsymbol{b}_i$. The optimization procedure identifies that the pixels that cause the most considerable decrease in confidence are flipped. That is, notionally

\begin{equation}
\arg \min_i F( \boldsymbol{x} + \boldsymbol{b}_i)      \text{ where }  \| \boldsymbol{\tilde{x}} - \boldsymbol{x}\|_0 \le k .
\end{equation}

\subsection{ECoBA: Efficient Combinatorial Black--box Adversarial Attack}
\label{ssec:shifting}

The ECoBA can be considered as a combination, in concert, of both additive and erosive perturbations. The errors and corresponding pixel numbers are stored in $D_{AP}$ and $D_{EP}$, merging them in a composite dictionary, $D_{AEP}$. For example, the top row of the $D_{AEP}$ contains the highest $\epsilon$ values for $\boldsymbol{w}_i$ and $\boldsymbol{b}_i$. For $k=1$, two pixels are flipped, corresponding to $\boldsymbol{w}_1$ and $\boldsymbol{b}_1$, resulting in no composite change in the number of black (or white) pixels. That is, there is no change in the $L_0$ norm. Accordingly, we utilize $k$ as the iteration index, corresponding to the number of flipped pixel pairs and the number of perturbations. 

The detailed steps of the proposed attack method are shown in algorithm \ref{alg:ecoba}.

\begin{algorithm}[H]
\caption{ECoBA}\label{alg:ecoba}
\begin{algorithmic}[1]
\Procedure{Adv}{$x$}\Comment{Create adversarial example of input image x}
\State $\tilde{x}\gets x$
\While{$ \arg \min_i F( \boldsymbol{x} + \boldsymbol{w}_i)  \text{ where } \| \boldsymbol{\tilde{x}} - \boldsymbol{x}\|_0 \le k$}
\State $w_i\gets \arg \max_i F(\boldsymbol{\tilde{x}) }$
\State $\epsilon_i \gets \boldsymbol F({x}_i) - \boldsymbol{F(\tilde{x}_i)}$
\State $D_{AP'}\gets w_i, \epsilon_i$\Comment{Dictionary with pixel index and its corresponding error}
\EndWhile
\State $D_{AP}\gets sort(D_{AP'})$\Comment{Sort the index of pixels starting from max error.}
\While{$ \arg \min_i F( \boldsymbol{x} + \boldsymbol{b}_i)  \text{ where } \| \boldsymbol{\tilde{x}} - \boldsymbol{x}\|_0 \le k$}
\State $b_i\gets \arg \max_i F(\boldsymbol{\tilde{x}) }$
\State $\epsilon_i \gets \boldsymbol F({x}_i) - \boldsymbol{F(\tilde{x}_i)}$
\State $D_{EP'}\gets b_i,\epsilon_i $
\EndWhile\label{euclidendwhile}
\State $D_{EP}\gets sort(D_{EP'})$\Comment{Sort the index of pixels starting from max error.}
\State $D_{AEP} \gets stack(D_{AP},D_{EP})$\Comment{Merge dictionaries into one.}
\State $\tilde{x}\gets \boldsymbol{x} + \boldsymbol{D_{{AEP}_i}}$\Comment{add perturbation couples from the merged dictionary}
\State \textbf{return} $\tilde{x}$
\EndProcedure
\end{algorithmic}
\end{algorithm}

\begin{figure}[H]
\begin{minipage}[b]{1.0\linewidth}
  \centering
  \centerline{\includegraphics[width=0.9\textwidth]{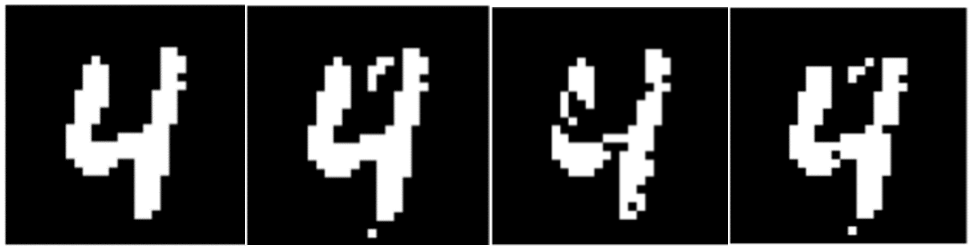}}
  \centerline{}\medskip
\end{minipage}
\caption{From left to right: original binary image, adversarial examples after only additive perturbations, only erosive perturbations, and final adversarial example with the proposed method.}
\label{fig:attack_tog}
\end{figure}

The amount of perturbations is controlled by $k$, which will be the step size in the simulations. Figure~\ref{fig:attack_tog} shows an example of the input image and the effect of perturbations.  

\section{SIMULATIONS}
\label{sec:sims}

We present simulations over two data sets and three different neural network-based classifiers in order to obtain comprehensive performance evaluations of the attack algorithms. Since the majority of optical characters involve with numbers and letters, we chose one data set for handwritten digits and another data set for handwritten letters. 

\subsection{Datasets}
\label{ssec:dataset}

Models were trained and tested on the hand--written digits MNIST \cite{lecun2010mnist} and letters EMNIST \cite{emnist} data sets. Images in the data sets are normalized between 0 and 1 as grayscale images are binarized using a global thresholding method with the threshold of 0.5. Both data sets consist of $28\times 28$ pixel images. MNIST and EMNIST have 70,000 and 145,000 examples, respectively. We use the split of 85\%-15\% of each dataset for training and testing.

\subsection{Models}
\label{ssec:models}

Three classifiers are employed for the training and testing. The simplest classifier, MLP-2, consists of only two fully connected layers with 128 and 64 nodes, respectively. The second classifier, a neural network architecture, is LeNet \cite{lenet}. Finally, the third classifier is a two-layer convolutional neural network (CNN), with 16 and 32 convolution filters of kernel size $5\times 5$. Training accuracies of each model on both datasets are shown in Table~\ref{tab:table-name}.

\begin{table}
\caption{\label{tab:table-name}Training performance of models.}
\centering
\begin{tabular}{ ||p{2cm}||p{2cm}|p{2cm}|p{2cm}||  }
 \hline
 \hline 
 \multicolumn{4}{||c||}{Top-1 Training Accuracy} \\
 \hline 
 \hline
 Dataset   & \multicolumn{1}{c|}{MLP2}  & \multicolumn{1}{c|}{LENET}  & \multicolumn{1}{c||}{CNN} \\
 \hline
 MNIST   & \multicolumn{1}{c|}{0.97 }   & \multicolumn{1}{c|}{0.99}    & \multicolumn{1}{c||}{0.99}\\
 \hline
 EMNIST  & \multicolumn{1}{c|}{0.91 }    & \multicolumn{1}{c|}{0.941}   & \multicolumn{1}{c||}{0.96}\\
 \hline
 \hline
\end{tabular}
\end{table}
The highest training accuracy was obtained with the CNN classifier, then LeNet and MLP-2, respectively. Training accuracies for both datasets with all classifiers are high enough to evaluate with testing samples. 
\subsection{Results}
\label{ssec:results}
\begin{figure}[H]
\subfloat{\includegraphics[width = .5\textwidth]{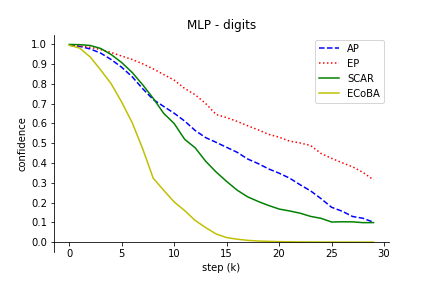}} 
\subfloat{\includegraphics[width = .5\textwidth]{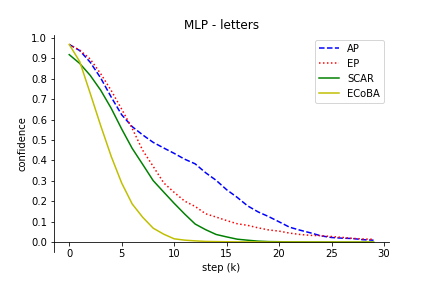}}\\
\subfloat{\includegraphics[width = .5\textwidth]{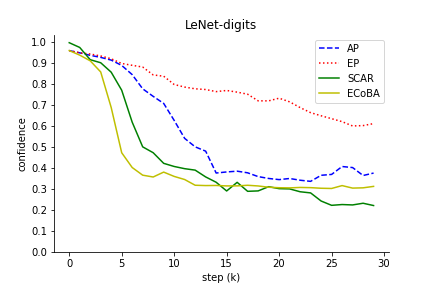}}
\subfloat{\includegraphics[width = .5\textwidth]{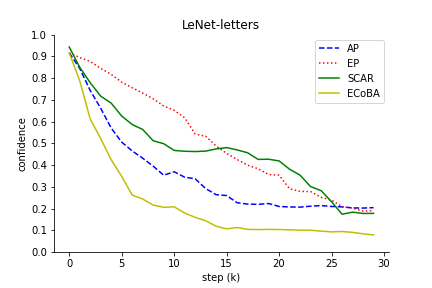}}\\
\subfloat{\includegraphics[width = .5\textwidth]{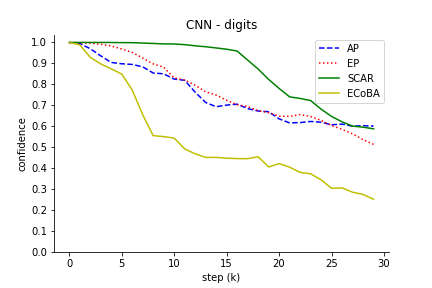}}
\subfloat{\includegraphics[width = .5\textwidth]{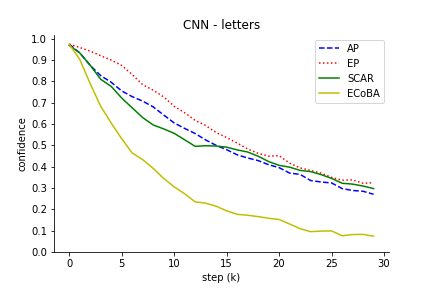}} 
\caption{Classification performance of input image with increasing step size. We include AP and EP as individual attack on input images to observe their effectiveness. }
\label{fig:compare}
\end{figure}

We evaluate the results of the proposed attacking method on three different neural network architectures over two different datasets. Figure~\ref{fig:compare} shows the attack performance over images from MNIST and EMNIST data sets. Ten input images are selected among correctly classified samples for the attack. The $Y$--axis of plots represents the averaged classification accuracy of input images, while the $X$--axis represents the number of iterations (number of added, removed, or shifted pixels). 

An observation of Figure~\ref{fig:compare} shows that all approaches yield successful attacks, with the proposed method generating the most successful attacks in all cases. Moreover, the classifier results on the adversarial examples yield very high confidence levels. The attack perturbations are applied even after the classifier gives the wrong label as a classification result to observe the attack strength. For instance, obtained average step size of ECoBA for misleading the MLP-2 classifier on the digit dataset is six. This means that changing the six pixels of the input image was enough to mislead the classifier. The averaged confidence level of the ground truth labels drops to zero when the attack perturbations are intensified on MLP-2. On the other hand, the proposed method generated more perturbations to mislead CNN classifier. We show the average step sizes for a successful attack for different attack types with respect to data sets in Table~\ref{tab:step-size}.

\begin{table}
\caption{\label{tab:step-size}Step sizes for a successful attack with respect to different classifiers.}
\centering
\begin{tabular}{ ||p{3cm}||p{1.5cm}|p{1.5cm}|p{1.5cm}|p{2cm}||  }
 \hline
 \hline
 \multicolumn{5}{||c||}{Average step sizes for a successful attack} \\
 \hline  
 classifier/method   & \multicolumn{1}{c|}{AP}  & \multicolumn{1}{c|}{EP}  & \multicolumn{1}{c|}{scar\cite{scar}}  & \multicolumn{1}{c||}{ECoBA}  \\
 \hline
 MLP(digits)   & \multicolumn{1}{c|}{9}    &\multicolumn{1}{c|}{11}      &\multicolumn{1}{c|}{8}   &\multicolumn{1}{c||}{\textbf{6}}  \\
 \hline
 MLP(letters)  & \multicolumn{1}{c|}{9}   & \multicolumn{1}{c|}{9}     & \multicolumn{1}{c|}{7}    & \multicolumn{1}{c||}{\textbf{5}} \\
 \hline
 LeNet(digits) & \multicolumn{1}{c|}{10}   & \multicolumn{1}{c|}{18}   & \multicolumn{1}{c|}{8}    & \multicolumn{1}{c||}{\textbf{6}}\\
 \hline
 LeNet(letters)& \multicolumn{1}{c|}{8}   & \multicolumn{1}{c|}{12}    & \multicolumn{1}{c|}{10}    & \multicolumn{1}{c||}{\textbf{5}} \\
 \hline
 CNN(digits)   & \multicolumn{1}{c|}{12}   & \multicolumn{1}{c|}{13}     & \multicolumn{1}{c|}{17}   & \multicolumn{1}{c||}{\textbf{8}} \\
 \hline
 CNN(letters)  & \multicolumn{1}{c|}{9}    & \multicolumn{1}{c|}{11}     & \multicolumn{1}{c|}{9}    & \multicolumn{1}{c||}{\textbf{7}}  \\
 \hline
 \hline 
\end{tabular}
\end{table}

\begin{figure}[H]

\begin{minipage}[b]{1.0\linewidth}
  \centering
  \centerline{\includegraphics[width=0.7\textwidth]{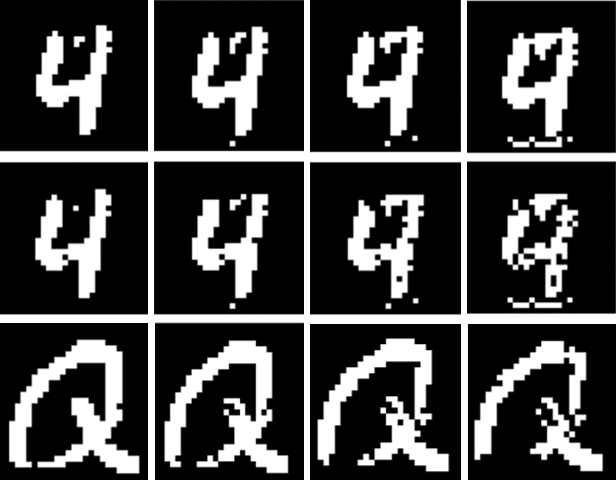}}
  \centerline{}\medskip
\end{minipage}

\caption{Class interpolation with increasing $k$. The first row: only AP, the second row: ECoBA, the last row: EP. }
\label{fig:interp}

\end{figure}

Another important outcome of the simulations is an observation of interpolations between classes, as reported earlier in \cite{tsipras2019robustness}. As we increase the number of iterations, the original input image interpolates to the closest class. Figure~\ref{fig:interp} provides an example of class interpolation. In this particular example, the ground truth label of the input image is four for the digit and Q for the letter. Once the attack intensifies, the input image is classified as digit nine and letter P, while the image evolves visually.

\section{Conclusion}
\label{sec:conclusion}
In this paper, we proposed an adversarial attack method on binary image classifiers in black-box settings, namely Efficient Combinatorial Black-box Adversarial Attack (ECoBA). We showed the inefficiency of most benchmark adversarial attack methods in binary image settings. Simulations show that the simplicity of the proposed method has enabled a strong adversarial attack with few perturbations. We showed the efficiency of the attack algorithm on two different data sets, MNIST and EMNIST. Simulations utilizing the MLP2, LENET, and CNN networks, show that even a small number of perturbations are enough to mislead classifiers with very high confidence.\\

%
%
%
%

\let\Origclearpage\clearpage
\let\clearpage\relax
\renewcommand{\bibname}{\protect\leftline{References}}
\bibliography{refs}

\begin{thebibliography}{20}
\providecommand{\natexlab}[1]{#1}
\providecommand{\url}[1]{\texttt{#1}}
\expandafter\ifx\csname urlstyle\endcsname\relax
  \providecommand{\doi}[1]{doi: #1}\else
  \providecommand{\doi}{doi: \begingroup \urlstyle{rm}\Url}\fi

\bibitem[Dalvi et~al.(2004)Dalvi, Domingos, Mausam, Sanghai, and Verma]{dalvi}
Nilesh Dalvi, Pedro Domingos, Mausam, Sumit Sanghai, and Deepak Verma.
\newblock Adversarial classification.
\newblock In \emph{Proceedings of the Tenth ACM SIGKDD International Conference
  on Knowledge Discovery and Data Mining}, KDD '04, page 99–108, New York,
  NY, USA, 2004. Association for Computing Machinery.
\newblock ISBN 1581138881.
\newblock \doi{10.1145/1014052.1014066}.
\newblock URL \url{https://doi.org/10.1145/1014052.1014066}.

\bibitem[Biggio and Roli(2018)]{Biggio_2018}
Battista Biggio and Fabio Roli.
\newblock Wild patterns: Ten years after the rise of adversarial machine
  learning.
\newblock \emph{Pattern Recognition}, 84:\penalty0 317–331, Dec 2018.
\newblock ISSN 0031-3203.
\newblock \doi{10.1016/j.patcog.2018.07.023}.
\newblock URL \url{http://dx.doi.org/10.1016/j.patcog.2018.07.023}.

\bibitem[Szegedy et~al.(2014)Szegedy, Zaremba, Sutskever, Bruna, Erhan,
  Goodfellow, and Fergus]{szegedy2014intriguing}
Christian Szegedy, Wojciech Zaremba, Ilya Sutskever, Joan Bruna, Dumitru Erhan,
  Ian Goodfellow, and Rob Fergus.
\newblock Intriguing properties of neural networks, 2014.

\bibitem[Goodfellow et~al.(2015)Goodfellow, Shlens, and Szegedy]{harnesing}
Ian Goodfellow, Jonathon Shlens, and Christian Szegedy.
\newblock Explaining and harnessing adversarial examples.
\newblock In \emph{International Conference on Learning Representations}, 2015.
\newblock URL \url{http://arxiv.org/abs/1412.6572}.

\bibitem[Carlini and Wagner(2017)]{carlini17}
Nicholas Carlini and David Wagner.
\newblock Towards evaluating the robustness of neural networks.
\newblock In \emph{2017 IEEE Symposium on Security and Privacy (SP)}, pages
  39--57, 2017.
\newblock \doi{10.1109/SP.2017.49}.

\bibitem[Nguyen et~al.(2015)Nguyen, Yosinski, and Clune]{nguyen2015deep}
Anh Nguyen, Jason Yosinski, and Jeff Clune.
\newblock Deep neural networks are easily fooled: High confidence predictions
  for unrecognizable images, 2015.

\bibitem[Moosavi-Dezfooli et~al.(2016)Moosavi-Dezfooli, Fawzi, and
  Frossard]{moosavidezfooli2016deepfool}
Seyed-Mohsen Moosavi-Dezfooli, Alhussein Fawzi, and Pascal Frossard.
\newblock Deepfool: a simple and accurate method to fool deep neural networks,
  2016.

\bibitem[Sharif et~al.(2016)Sharif, Bhagavatula, Bauer, and Reiter]{sharif}
Mahmood Sharif, Sruti Bhagavatula, Lujo Bauer, and Michael~K. Reiter.
\newblock Accessorize to a crime: Real and stealthy attacks on state-of-the-art
  face recognition.
\newblock In \emph{Proceedings of the 2016 ACM SIGSAC Conference on Computer
  and Communications Security}, CCS '16, page 1528–1540, New York, NY, USA,
  2016. Association for Computing Machinery.
\newblock ISBN 9781450341394.
\newblock \doi{10.1145/2976749.2978392}.
\newblock URL \url{https://doi.org/10.1145/2976749.2978392}.

\bibitem[Kurakin et~al.(2016)Kurakin, Goodfellow, and Bengio]{kurakin16}
Alexey Kurakin, Ian~J. Goodfellow, and Samy Bengio.
\newblock Adversarial examples in the physical world.
\newblock \emph{CoRR}, abs/1607.02533, 2016.
\newblock URL \url{http://arxiv.org/abs/1607.02533}.

\bibitem[Eykholt et~al.(2018)Eykholt, Evtimov, Fernandes, Li, Rahmati, Xiao,
  Prakash, Kohno, and Song]{eykholt2018robust}
Kevin Eykholt, Ivan Evtimov, Earlence Fernandes, Bo~Li, Amir Rahmati, Chaowei
  Xiao, Atul Prakash, Tadayoshi Kohno, and Dawn Song.
\newblock Robust physical-world attacks on deep learning models, 2018.

\bibitem[Smith(2007)]{tesseractsrev07}
Ray Smith.
\newblock An overview of the tesseract ocr engine.
\newblock In \emph{Proc. Ninth Int. Conference on Document Analysis and
  Recognition (ICDAR)}, pages 629--633, 2007.

\bibitem[Madry et~al.(2019)Madry, Makelov, Schmidt, Tsipras, and
  Vladu]{madry2019deep}
Aleksander Madry, Aleksandar Makelov, Ludwig Schmidt, Dimitris Tsipras, and
  Adrian Vladu.
\newblock Towards deep learning models resistant to adversarial attacks, 2019.

\bibitem[Su et~al.(2019)Su, Vargas, and Sakurai]{onepixel}
Jiawei Su, Danilo~Vasconcellos Vargas, and Kouichi Sakurai.
\newblock One pixel attack for fooling deep neural networks.
\newblock \emph{IEEE Transactions on Evolutionary Computation}, 23\penalty0
  (5):\penalty0 828–841, Oct 2019.
\newblock ISSN 1941-0026.
\newblock \doi{10.1109/tevc.2019.2890858}.
\newblock URL \url{http://dx.doi.org/10.1109/TEVC.2019.2890858}.

\bibitem[Tramèr et~al.(2017)Tramèr, Papernot, Goodfellow, Boneh, and
  McDaniel]{transfer}
Florian Tramèr, Nicolas Papernot, Ian Goodfellow, Dan Boneh, and Patrick
  McDaniel.
\newblock The space of transferable adversarial examples, 2017.

\bibitem[Balkanski et~al.(2020)Balkanski, Chase, Oshiba, Rilee, Singer, and
  Wang]{scar}
Eric Balkanski, Harrison Chase, Kojin Oshiba, Alexander Rilee, Yaron Singer,
  and Richard Wang.
\newblock Adversarial attacks on binary image recognition systems, 2020.

\bibitem[Wang et~al.(2021)Wang, Zhang, Shen, Yu, and
  Wang]{binarization_deffense}
Yutong Wang, Wenwen Zhang, Tianyu Shen, Hui Yu, and Fei{-}Yue Wang.
\newblock Binary thresholding defense against adversarial attacks.
\newblock \emph{Neurocomputing}, 445:\penalty0 61--71, 2021.
\newblock \doi{10.1016/j.neucom.2021.03.036}.
\newblock URL \url{https://doi.org/10.1016/j.neucom.2021.03.036}.

\bibitem[LeCun et~al.(2010)LeCun, Cortes, and Burges]{lecun2010mnist}
Yann LeCun, Corinna Cortes, and CJ~Burges.
\newblock Mnist handwritten digit database.
\newblock \emph{ATT Labs [Online]. Available:
  http://yann.lecun.com/exdb/mnist}, 2, 2010.

\bibitem[Cohen et~al.(2017)Cohen, Afshar, Tapson, and van Schaik]{emnist}
Gregory Cohen, Saeed Afshar, Jonathan Tapson, and Andr{\'e} van Schaik.
\newblock Emnist: an extension of mnist to handwritten letters.
\newblock \emph{arXiv preprint arXiv:1702.05373}, 2017.

\bibitem[Lecun et~al.(1998)Lecun, Bottou, Bengio, and Haffner]{lenet}
Y.~Lecun, L.~Bottou, Y.~Bengio, and P.~Haffner.
\newblock Gradient-based learning applied to document recognition.
\newblock \emph{Proceedings of the IEEE}, 86\penalty0 (11):\penalty0
  2278--2324, 1998.
\newblock \doi{10.1109/5.726791}.

\bibitem[Tsipras et~al.(2019)Tsipras, Santurkar, Engstrom, Turner, and
  Madry]{tsipras2019robustness}
Dimitris Tsipras, Shibani Santurkar, Logan Engstrom, Alexander Turner, and
  Aleksander Madry.
\newblock Robustness may be at odds with accuracy, 2019.

\end{thebibliography}
\let\clearpage\Origclearpage
\end{document}